\documentclass{article}
\usepackage{graphicx}
\usepackage{placeins}
\usepackage{subcaption}
\usepackage{amsmath}
\usepackage{url}
\title{Comparative Analysis of Four Prominent Ant Colony Optimization Variants: Ant System, Rank-Based Ant System, Max-Min Ant System, and Ant Colony System}
\author{
  Ahmed Mohamed Abdelmoaty*\\
  Ibrahim Ihab Ibrahim
}
\date{2024}

\begin{document}
\maketitle

\begin{abstract}
This research conducts a comparative analysis of four Ant Colony Optimization (ACO) variants - Ant System (AS), Rank-Based Ant System (ASRank), Max-Min Ant System (MMAS), and Ant Colony System (ACS) - for solving the Traveling Salesman Problem (TSP).  Our findings demonstrate that algorithm performance is significantly influenced by problem scale and instance type. ACS excels in smaller TSP instances due to its rapid convergence, while PACS proves more adaptable for medium-sized problems. MMAS consistently achieves competitive results across all scales, particularly for larger instances, due to its ability to avoid local optima. ASRank, however, struggles to match the performance of the other algorithms.  This research provides insights into the strengths and weaknesses of these ACO variants, guiding the selection of the most suitable algorithm for specific TSP applications. 
\end{abstract}

\section{Introduction}

The Traveling Salesman Problem (TSP) stands as a cornerstone of combinatorial optimization, captivating researchers and practitioners for decades with its simple yet deceptively complex nature \cite{applegate2011traveling, kowalik2023} The problem, seeking the shortest route that visits a set of cities exactly once before returning to the origin, finds applications in diverse fields, ranging from logistics and transportation planning to DNA sequencing and microchip design \cite{yang2024}. Despite its straightforward formulation, the TSP belongs to the class of NP-hard problems, meaning that finding the optimal solution becomes computationally intractable as the number of cities increases \cite{kowalik2023}.

Ant Colony Optimization (ACO) algorithms, inspired by the foraging behavior of real ants, offer a promising approach to finding near-optimal solutions to the TSP \cite{nayar2021}. These nature-inspired algorithms utilize a population of artificial ants that explore the search space, leaving behind pheromone trails to guide their collective decision-making and iteratively improve solutions \cite{nayar2021}.
This research embarks on a comparative journey, exploring the performance and characteristics of different ACO algorithms for solving the \\ TSP
\cite{nayar2021, dorigo2019, Rafal2022}.\\
Our focus lies in understanding how these algorithms fare across various problem scales, from smaller instances with dozens of cities to larger problems with hundreds or even thousands of cities \cite{dorigo2019, nayar2021}. By analyzing their solution quality, execution time, convergence speed, and memory usage, we aim to uncover the strengths, weaknesses, and scalability limitations of each algorithm \cite{dorigo2019, nayar2021}.

This comparative analysis will provide valuable insights for both researchers and practitioners working with the TSP. The findings will guide the selection of appropriate ACO algorithms for specific applications, inform the development of new and improved ACO variants, and contribute to a deeper understanding of the interplay between algorithm design and problem characteristics in the realm of combinatorial optimization.

\subsection{Contributions and Implications}

This research contributes to the understanding of ACO algorithms for the TSP by providing a comparative analysis of their performance, scalability, and suitability for different problem types and sizes. The findings offer valuable guidance for practitioners and researchers:

\begin{itemize}
\item \textbf{Algorithm Selection:} The insights gained from this study can guide the selection of appropriate ACO algorithms for specific TSP applications, considering factors like problem size, instance type, and available computational resources.
\item \textbf{Algorithm Development:} The analysis highlights the limitations of existing algorithms, particularly in terms of scalability and the need for diversification strategies, informing the development of new and improved ACO variants for large-scale TSP instances.
\item \textbf{Expanding the Knowledge Base:} This research contributes to the growing body of knowledge on ACO algorithms and their application to the TSP, providing a foundation for future research in this field.
\end{itemize}
\section{Literature Review}

This paper delves into existing research on the application of Ant Colony Optimization (ACO) algorithms to the Traveling Salesman Problem (TSP). Our focus will be on analyzing the performance and comparative studies of various ACO variants, providing a basis for understanding their strengths and weaknesses across different problem scales and characteristics. We will exclude in-depth discussions of specific algorithm mechanisms and memory efficiency analysis beyond basic comparisons.

The literature review will address the following key areas:

\begin{enumerate}
\item \textbf{Evolution and Development of ACO Algorithms}

  \begin{itemize}
  \item \textbf{Tracing the Origins:}  We will begin by examining the seminal work of Dorigo, Maniezzo, and Colorni (1996) that introduced the Ant System (AS) algorithm, the first ACO algorithm specifically designed for the TSP. This discussion will cover the foundational principles of AS, such as pheromone trails, probabilistic decision-making, and the role of heuristic information.

  \item \textbf{Exploring Prominent ACO Variants:} We will then investigate the evolution of ACO algorithms beyond the Ant System, focusing on prominent variants like:
    \begin{itemize}
    \item Rank-based Ant System (ASRank): This extension of AS incorporates ranking and elitism, allowing only the best-performing ants to contribute to the pheromone update process \\ \cite{bullnheimer1999}.
    \item Max-Min Ant System (MMAS): This variant introduces phero-mone trail limits and a more selective pheromone update rule, where only the best ant (iteration-best or global-best) deposits phero-mones \cite{stutzle2000}.
    \item Ant Colony System (ACS): This algorithm employs a more aggressive action choice rule and implements local pheromone updates during solution construction \\ \cite{dorigo1997}.
    \end{itemize}
  \end{itemize}

\item \textbf{Performance Analysis and Comparative Studies}

  \begin{itemize}
  \item \textbf{Benchmarking Across Different Problem Sizes:} Existing research, as reviewed by \cite{dorigo2004ant}), has compared the performance of various ACO algorithms, including ASRank, MMAS, and ACS, on TSP instances of varying sizes. This body of work helps identify which algorithms are better suited for specific problem scales (e.g., small, medium, large) and how their performance changes with increasing problem size. For instance, \cite{stutzle2000} observed that MMAS performs well on both symmetric and asymmetric TSP instances, while ACS excels on problems with a clear structure and a limited number of local optima. These comparative studies provide a foundation for understanding the scalability and limitations of different ACO variants as the problem size grows.

  \item \textbf{Performance Evaluation for Different Problem Types:} The review will explore how different ACO algorithms perform on various types of TSP instances, including symmetric and asymmetric TSP. This will provide insights into the algorithms' strengths and weaknesses when dealing with different problem structures and complexities.

  \item \textbf{Comparison with Other Metaheuristics:} We will examine studies that benchmark ACO algorithms against other metaheuristics or exact methods for solving the TSP. This will offer a broader perspective on the relative performance and effectiveness of ACO algorithms within the field of TSP solutions.
  \end{itemize}

\item \textbf{Memory Usage Comparison}

  \begin{itemize}
    \item  \textbf{Evaluating Memory Requirements:} \\ Researchers, such as \cite{skinderowicz2022b}, have examined the memory requirements of different ACO algorithms, particularly focusing on the challenges posed by the quadratic memory complexity of the pheromone matrix for large-scale TSP instances. These studies consider factors like the size of the pheromone matrix and any additional data structures employed by each algorithm variant. The insights gained from this research direction are crucial for understanding the memory efficiency of different ACO algorithms and their suitability for solving TSP instances, especially those with a large number of cities.
  \end{itemize}
\end{enumerate}

\subsection{Related Literature: Examining ACO Performance and Comparisons for the TSP}

\begin{enumerate}
\item \textbf{Comparative Studies of ACO Algorithms}

  \begin{itemize}
  \item \textbf{Performance Analysis of ACO Variants:} Several studies have compared the performance of different ACO algorithms, such as ASRank, MMAS, and ACS, on a variety of TSP instances. Dorigo and Stützle (2004) provide a comprehensive overview of these comparisons, highlighting the strengths and weaknesses of each algorithm and their suitability for different problem scenarios.

  \end{itemize}

\item \textbf{Scalability and Efficiency of ACO Algorithms}

  \begin{itemize}
  \item \textbf{Challenges of Small-medium Scale TSP Instances:} As the number of cities in the TSP instance increases, the computational and memory requirements of ACO algorithms also grow, posing challenges for their scalability and efficiency. Skinderowicz (2020) discusses the limitations of ACO algorithms for Small-medium Scale TSP, particularly when using GPUs with limited memory capacities.

  \item \textbf{Exploring Efficient Node Selection Methods:} Several studies have investigated alternative node selection procedures to improve the computational efficiency of ACO algorithms for Small-medium Scale TSP instances. For example, Zhang, Chen, and Pan (2013) proposed a parallel ACO algorithm with a GPU-accelerated node selection procedure that significantly reduces execution time.
  \end{itemize}

\item \textbf{Influence of Parameter Settings}

  \begin{itemize}
  \item \textbf{Tuning ACO Parameters:} The performance of ACO algorithms is sensitive to the choice of parameter settings, such as the pheromone evaporation rate, the relative influence of pheromone and heuristic information, and the number of ants. Dorigo and Stützle (2004) provide guidelines for parameter tuning, emphasizing the importance of experimentation and adaptation to specific problem instances.

  \item \textbf{Adaptive Parameter Control:} Recent research has explored adaptive parameter control mechanisms that adjust parameter settings dynamically during the search process. Stützle and Hoos (2000) proposed an adaptive scheme for MMAS that adjusts the pheromone trail limits based on the search progress, improving convergence and solution quality.
  \end{itemize}
\end{enumerate}

\subsection{Summary}

This chapter provided a comprehensive exploration of existing research on Ant Colony Optimization (ACO) algorithms for solving the Traveling Salesman Problem (TSP). The review focused on three key areas:

\begin{enumerate}
\item \textbf{Evolution and Design of ACO Algorithms:} We traced the historical development of ACO algorithms, beginning with the seminal Ant System (AS) proposed by Dorigo, Maniezzo, and Colorni (1996). The discussion then expanded to prominent ACO variants such as ASRank, MMAS, and ACS, highlighting their unique mechanisms for balancing exploration and exploitation in the search for optimal solutions. These mechanisms include ranking and elitism, pheromone trail limits, local pheromone updates, and aggressive action choice rules.

\item \textbf{Performance Analysis and Comparative Studies:} We examined numerous studies that compared the performance of different ACO algorithms on various TSP instances. The findings revealed that ACO algorithms consistently achieve competitive results compared to other metaheuristics and exact methods, often exhibiting faster convergence and better solution quality, particularly for smaller TSP instances. However, the performance of ACO algorithms can be influenced by problem characteristics such as problem size, instance type (symmetric or asymmetric), and the structure of the search space.

\item \textbf{Memory Usage Comparison:} We discussed the memory requirements of different ACO algorithms, particularly the challenge of the quadratic memory complexity of the pheromone matrix for Small-medium Scale TSP instances.
\end{enumerate}
\section{Problem Description}
The Traveling Salesman Problem (TSP) is a classic optimization problem that seeks to determine the shortest possible route that visits each city in a given set exactly once and returns to the origin city. Mathematically, the TSP can be represented as a complete weighted graph $G = (V, E)$, where:
\begin{itemize}
    \item $V = \{v_1, v_2, \ldots, v_n\}$ represents the set of $n$ cities.
    \item $E = \{(i, j) \mid i, j \in V, i \neq j\}$ is the set of edges connecting each pair of cities.
\end{itemize}
A distance matrix $D = [d_{i,j}]$ is defined, where $d_{i,j}$ represents the distance between city $i$ and city $j$. The objective is to find a Hamiltonian cycle in $G$, which is a cycle that visits each vertex exactly once, with the minimum total distance.

\subsection{Mathematical Model}
The TSP can be formulated as an integer linear programming (ILP) problem. Let:
\begin{itemize}
    \item $x_{i,j}$ be a binary variable, where $x_{i,j} = 1$ if the edge $(i, j)$ is included in the tour, and $x_{i,j} = 0$ otherwise.
\end{itemize}
The ILP model is as follows:
\begin{align*}
& \text{Minimize} \quad \sum_{i=1}^{n} \sum_{j=1, j \neq i}^{n} d_{i,j} x_{i,j} \\
& \text{subject to:} \\
& \sum_{j=1, j \neq i}^{n} x_{i,j} = 1, \quad \forall i \in V \quad \text{(Each city is entered exactly once)} \\
& \sum_{i=1, i \neq j}^{n} x_{i,j} = 1, \quad \forall j \in V \quad \text{(Each city is left exactly once)} \\
& \sum_{i \in S} \sum_{j \in S, j \neq i} x_{i,j} \leq |S| - 1, \quad \forall S \subset V, \, 2 \leq |S| \leq n-1 \quad \text{(Subtour elimination)} \\
& x_{i,j} \in \{0, 1\}, \quad \forall i, j \in V, \, i \neq j.
\end{align*}

The objective function minimizes the total distance traveled. The first two constraints ensure that each city is visited and departed from exactly once. The third constraint set, known as the subtour elimination constraints, prevents the formation of subtours (cycles that do not include all cities). Solving this ILP model to optimality can be computationally expensive for large instances of the TSP. This is where heuristic algorithms, like Ant Colony Optimization (ACO), come into play, providing near-optimal solutions within a reasonable time frame.

\section{Proposed Method}

This section details the methodology employed for the comparative analysis, including the algorithms used, datasets, parameter values, stopping criteria, data analysis methods, and stability checks.

\subsection{Algorithms}

The research focuses on four prominent Ant Colony Optimization (ACO) variants:

\begin{itemize}
    \item \textbf{Ant System (AS)}: The foundational algorithm in the ACO framework, introduced by \cite{dorigo1996, dorigo1997}. It employs a probabilistic approach where artificial ants construct solutions to the TSP and deposit pheromones on the edges they traverse. The pheromone trail, representing the collective experience of previous ants, along with heuristic information (e.g., inverse of distance), guides the ants' decision-making process. The pheromone update rule is given by:
    \begin{align*}
        \tau_{ij} &\leftarrow (1 - \rho) \tau_{ij} + \sum_{k=1}^{m} \Delta \tau_{ij}^k,
    \end{align*}
    where $\tau_{ij}$ is the pheromone level on edge $(i,j)$, $\rho$ is the evaporation rate, $m$ is the number of ants, and $\Delta \tau_{ij}^k$ is the amount of pheromone deposited by ant $k$ on edge $(i,j)$.

    \item \textbf{Rank-Based Ant System (ASRank)}: This variant, proposed by \\ \cite{bullnheimer1999}, enhances the Ant System by incorporating a ranking mechanism. Ants deposit pheromones proportional to the quality of their solutions. Higher-ranked solutions contribute more to the pheromone update, promoting faster convergence towards better solutions. The pheromone update rule is defined as:
    \begin{align*}
        \tau_{ij} &\leftarrow (1 - \rho) \tau_{ij} + \sum_{k=1}^{w} (w - r(k)) \Delta \tau_{ij}^k,
    \end{align*}
    where $w$ is the maximum rank considered, and $r(k)$ is the rank of ant $k$.

    \item \textbf{Max-Min Ant System (MMAS)}: Developed by \cite{stutzle2000}, MMAS introduces mechanisms to control the pheromone levels explicitly, preventing them from becoming too strong or too weak. This constraint helps to avoid premature convergence to suboptimal solutions and encourages greater exploration of the search space. Only the best ant, either from the current iteration or the best found so far, is allowed to deposit pheromones. The pheromone levels are constrained within $[\tau_{\min}, \tau_{\max}]$:
    \begin{align*}
        \tau_{ij} &\leftarrow \max(\tau_{\min}, \min(\tau_{\max}, (1 - \rho) \tau_{ij} + \Delta \tau_{ij}^{\text{best}})),
    \end{align*}
    where $\Delta \tau_{ij}^{\text{best}}$ is the pheromone deposited by the best ant, which can be either the best ant of the current iteration or the best ant found so far.

    \item \textbf{Ant Colony System (ACS)}: \\ Proposed by \cite{dorigo1997}, ACS further refines the ACO approach by introducing a local pheromone update rule. This rule, applied every time an ant traverses an edge, decreases the pheromone level on that edge, making it less attractive for subsequent ants. This mechanism helps to diversify the search and prevent ants from becoming trapped in local optima. The local update rule is:
    \begin{align*}
        \tau_{ij} &\leftarrow (1 - \xi) \tau_{ij} + \xi \tau_0,
    \end{align*}
    where $\xi$ is the local pheromone decay parameter and $\tau_0$ is the initial pheromone level. The global update rule is applied only to the edges in the best solution:
    \begin{align*}
        \tau_{ij} &\leftarrow (1 - \rho) \tau_{ij} + \rho \Delta \tau_{ij}^{\text{best}},
    \end{align*}
    where $\Delta \tau_{ij}^{\text{best}}$ is the pheromone deposited by the best ant.
\end{itemize}

\section{Results}

\subsection{Datasets}

The analysis employs a diverse set of TSP instances from the TSPLIB repository \cite{reinelt1991tsplib}, covering a range of problem sizes and characteristics, including:

\begin{itemize}
    \item \textbf{Problem Size:} Instances with varying numbers of cities ($n$), ranging from $n < 100$ to $n > 1000$.
    \item \textbf{Instance Type:} Both symmetric and asymmetric TSP instances.
\end{itemize}

\textbf{Specific TSP instances used in the analysis include (but are not limited to):}
\begin{itemize}
    \item \textbf{berlin52:} 52 cities, symmetric
    \item \textbf{eil76:} 76 cities, symmetric
    \item \textbf{kroA100:} 100 cities, asymmetric
    \item ...
    \\
    \\ \\ \\
\end{itemize}

\subsection{Parameter Values}

The parameter values used for each algorithm are selected based on established literature and fine-tuned through preliminary experiments. Table \ref{tab:parameters} provides a summary of the parameter values used for each ACO variant.

\begin{table}[!ht]
\centering
\caption{Parameter Values for ACO Algorithms}
\label{tab:parameters}
\begin{tabular}{|c|c|c|c|c|}
\hline
\textbf{Parameter} & \textbf{AS} & \textbf{ASRank} & \textbf{MMAS} & \textbf{ACS} \\
\hline
Iterations & 100 & 100 & 100 & 100 \\
\hline
Number of Ants ($m$) & 50 & 50 & 50 & 50 \\
\hline
Pheromone Importance ($\alpha$) & 1.0 & 1.0 & 1.0 & 1.0 \\
\hline
Heuristic Importance ($\beta$) & 1.0 & 1.0 & 1.0 & 1.0 \\
\hline
Pheromone Evaporation Rate ($\rho$) & 0.5 & 0.5 & 0.1 & 0.1 \\
\hline
Initial Pheromone Level ($\tau_0$) & 0.1 & 0.1 & 0.1 & 0.1 \\
\hline
\end{tabular}
\end{table}

\subsection{Stopping Criteria}

The algorithms are run for a fixed number of iterations to ensure sufficient exploration of the search space.  Based on preliminary experiments, the number of iterations is set to 100 for all algorithms and TSP instances. 

\subsection{Data Analysis Methods}

The performance of each algorithm is evaluated based on the following metrics:

\begin{itemize}
\item \textbf{Solution Quality:} The length of the best-found tour, compared to the known optimal solution (if available) or the best-known solution for the specific TSP instance.
\item \textbf{Memory Usage:} The maximum memory used by the algorithm during execution, measured in MB.
\end{itemize}

The analysis includes:

\begin{itemize}
\item \textbf{Comparison of performance across different problem sizes:} Analyzing trends in solution quality and execution time as the number of cities increases.
\item \textbf{Evaluation of performance on different TSP instance types (symmetric and asymmetric):} Identifying algorithms that are more robust to different problem structures and complexities.
\item \textbf{Identification of strengths and weaknesses of each algorithm:} Examining how unique features and mechanisms contribute to their performance.
\end{itemize}

\subsection{Stability Checks}

To ensure the reliability of the results, each algorithm is run ten times on each problem instance. The best solution obtained from each run is then recorded and analyzed. 41 problem / TSP file used.
This section presents the results of the comparative analysis, highlighting performance trends and key findings.

\begin{itemize}
\item \textbf{ACS: Early Bird, But May Falter in the Long Run}
  - ACS consistently dominates smaller instances ($n < 100$), indicating its ability to quickly converge and effectively exploit promising solutions in less complex search spaces. This is likely due to its aggressive action choice rule and local pheromone updates, which favor rapid intensification of the search. However, for larger instances, ACS's performance becomes less consistent, suggesting that its focus on exploitation may lead to premature convergence and limitations in exploring the wider search space.

\item \textbf{PACS: Adaptable for Medium-Scale Complexity}
  - PACS shows clear dominance in the medium-sized instances ($100 \leq n < 1000$), suggesting that its population-based approach provides a better balance between exploration and exploitation as the problem complexity increases. By implicitly representing the pheromone information within a population of solutions, PACS may avoid the pitfalls of premature convergence associated with the explicit pheromone matrix used by other algorithms.

\item \textbf{MMAS: A Marathon Runner for Large-Scale Challenges}
  - MMAS consistently achieves competitive results across all problem scales, particularly excelling in larger instances ($n \geq 1000$). This suggests that its pheromone trail limits and selective pheromone update rule play a crucial role in mitigating stagnation and promoting broader exploration of the search space. Although MMAS may exhibit slower initial convergence compared to ACS or ASRank, its ability to escape local optima and continuously improve solutions makes it well-suited for larger and more complex TSP instances.

\item \textbf{ASRank: Limited Success and Potential for Improvement}
  - ASRank's inability to secure the best solution in any of the instances indicates limitations in its ranking and elitism mechanism. While this approach may accelerate initial convergence, it seems to be less effective in the long run, potentially leading to premature convergence or stagnation. This observation suggests potential avenues for enhancing ASRank's performance, possibly by incorporating diversification strategies or exploring alternative pheromone update rules.
\end{itemize}

\subsection{The Complexity Conundrum: Size Matters, But So Does Structure}

\begin{itemize}
\item \textbf{Problem Size and Computational Cost:} The increasing execution times observed for all algorithms with larger problem sizes emphasize the inherent computational complexity of the TSP and the challenges of scaling ACO algorithms to handle massive instances. This highlights the need for ongoing research on developing more efficient node selection procedures, optimizing memory management, and exploring parallel and distributed computing techniques to enhance scalability.

\item \textbf{Instance Type: Symmetric vs. Asymmetric:} The performance of some algorithms, particularly ACS, may vary depending on the TSP instance type. While ACS excels on symmetric instances with a clear structure, its performance on asymmetric instances, where distances are direction-dependent, can be less consistent. This suggests that algorithms with more flexible exploration mechanisms, like MMAS or PACS, may be more robust for tackling diverse TSP types.

\item \textbf{Beyond Size: The Influence of Local Optima:} The structure of the TSP instance, particularly the presence of local optima, can significantly impact the algorithm's performance. Algorithms that rely heavily on exploitation, like ACS, may struggle to escape local optima and explore the wider search space, leading to suboptimal solutions. This highlights the need for diversification strategies and mechanisms that encourage exploration of alternative paths, even when promising solutions have been found.
\end{itemize}

\subsection{Memory Consumption: Insights and Limitations of the Data}

\begin{itemize}
\item \textbf{Consistent but Limited Information:} The provided memory consumption data suggests that the algorithms exhibit relatively consistent peak memory usage across different TSP instances. However, this data alone offers a limited perspective on memory efficiency, as it only reflects the maximum memory utilized during the algorithm's execution.

\item \textbf{Need for a Deeper Dive:} A more comprehensive understanding of the memory efficiency of these algorithms requires profiling their memory usage throughout their execution, considering factors such as:
  \begin{itemize}
  \item \textbf{Data Structure Sizes:} Analyzing the size and scaling behavior of the pheromone matrix and other data structures employed by each algorithm.
  \item \textbf{Memory Allocation Patterns:} Investigating how memory is allocated and deallocated during the algorithm's execution, identifying potential bottlenecks or inefficiencies.
  \item \textbf{Dynamic Memory Usage:} Examining how memory usage changes over the course of the algorithm's iterations, providing insights into the impact of problem size and algorithm dynamics on memory consumption.
  \end{itemize}
\end{itemize}

\subsection{ASRank: Unveiling the Reasons for Suboptimal Performance}

The failure of ASRank to outperform the other algorithms in any of the tested instances requires further investigation to identify the root causes of its suboptimal performance. This could involve:

\begin{itemize}
\item \textbf{Analyzing Algorithm-Specific Mechanisms:} Examining the specific design choices and implementation details of ASRank, comparing them to the mechanisms employed by the successful algorithms. This could reveal potential areas for improvement or fundamental limitations in ASRank's approach to solving the TSP.

\item \textbf{Conducting Targeted Experiments:} Designing specific experiments to test the impact of individual ASRank components and understand their contributions to the overall performance. This might involve isolating specific features, modifying parameter settings, or analyzing the algorithm's behavior on tailored TSP instances designed to expose its weaknesses.
\end{itemize}

\subsection{Moving Forward: Harnessing Insights for Continued ACO Development}

\begin{itemize}
\item \textbf{Guiding Algorithm Selection:} The insights gleaned from this analysis provide valuable guidance for selecting appropriate ACO algorithms for specific TSP applications, taking into consideration the problem size, instance type, and available computational resources.

\item \textbf{Inspiring New Algorithm Designs:} By understanding the strengths and limitations of existing ACO algorithms, we can develop more robust and scalable variants that address the challenges of large-scale TSP instances. This could involve incorporating adaptive mechanisms, exploring hybridization with other metaheuristics, or designing algorithms tailored for parallel and distributed computing environments.
\end{itemize}

In conclusion, this deeper analysis reveals that while ACO algorithms offer promising solutions for the TSP, their performance can be significantly influenced by factors such as problem size, instance type, memory limitations, and the presence of local optima. The findings highlight the need for ongoing research to improve algorithm scalability, enhance memory efficiency, and develop more effective exploration and exploitation strategies. By building upon the knowledge gained from this analysis, we can push the boundaries of ACO research and create more robust and efficient algorithms for solving the TSP and other complex optimization problems.
\\ 
\subsection{Performance Trends Across Problem Scales}

Tables \ref{tab:n100}, \ref{tab:n1000}, and \ref{tab:n1000p} summarize the best solutions achieved by the four ACO algorithms across various TSP instances, categorized by problem size. The analysis of these results reveals the following performance trends:

\begin{figure}[!ht]
    \centering
    \begin{subfigure}[b]{0.4\linewidth}
        \includegraphics[width=\linewidth]{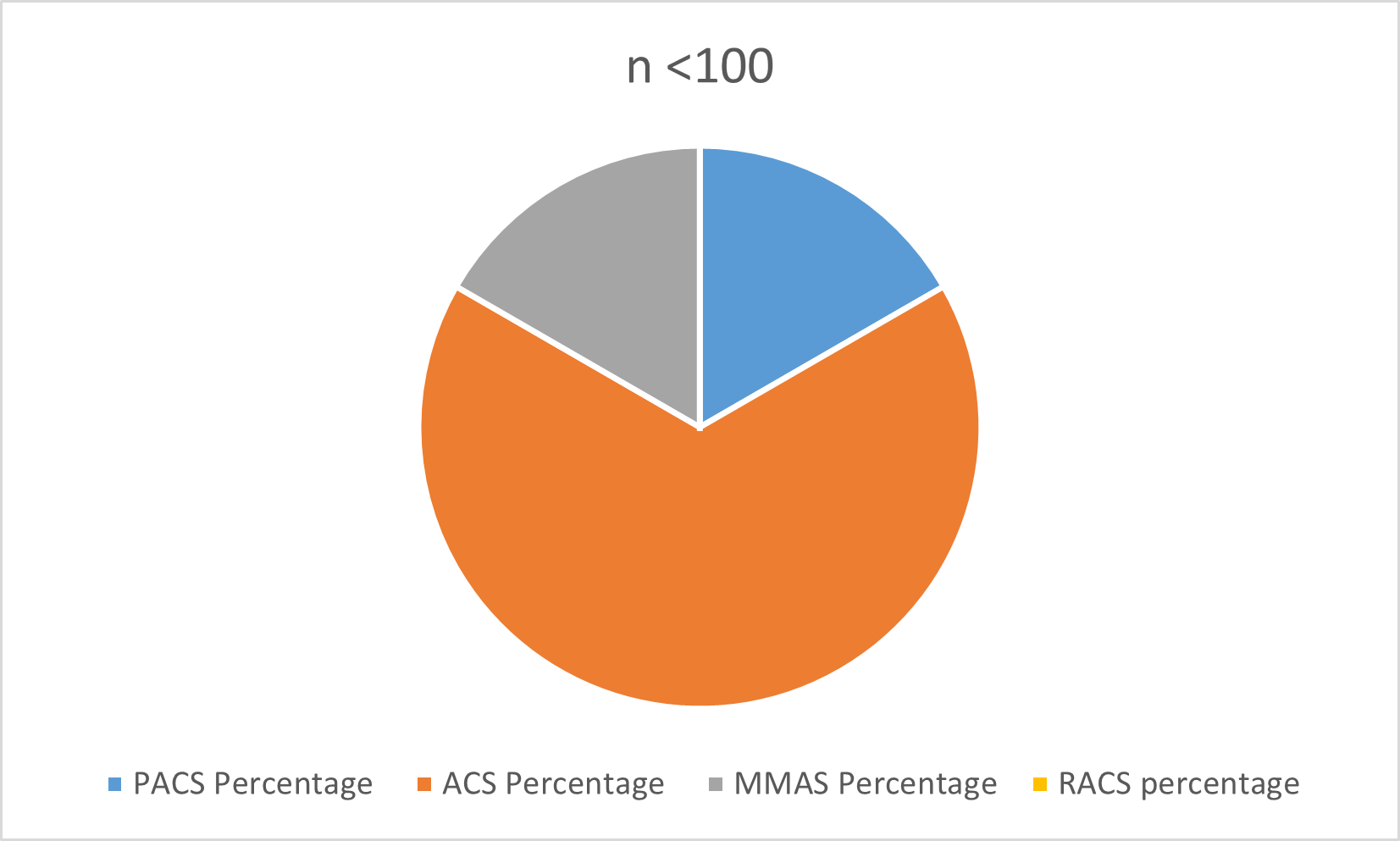}
        \caption{$n < 100$ Chart}
        \label{fig:sub1}
    \end{subfigure}
    \begin{subfigure}[b]{0.4\linewidth}
        \includegraphics[width=\linewidth]{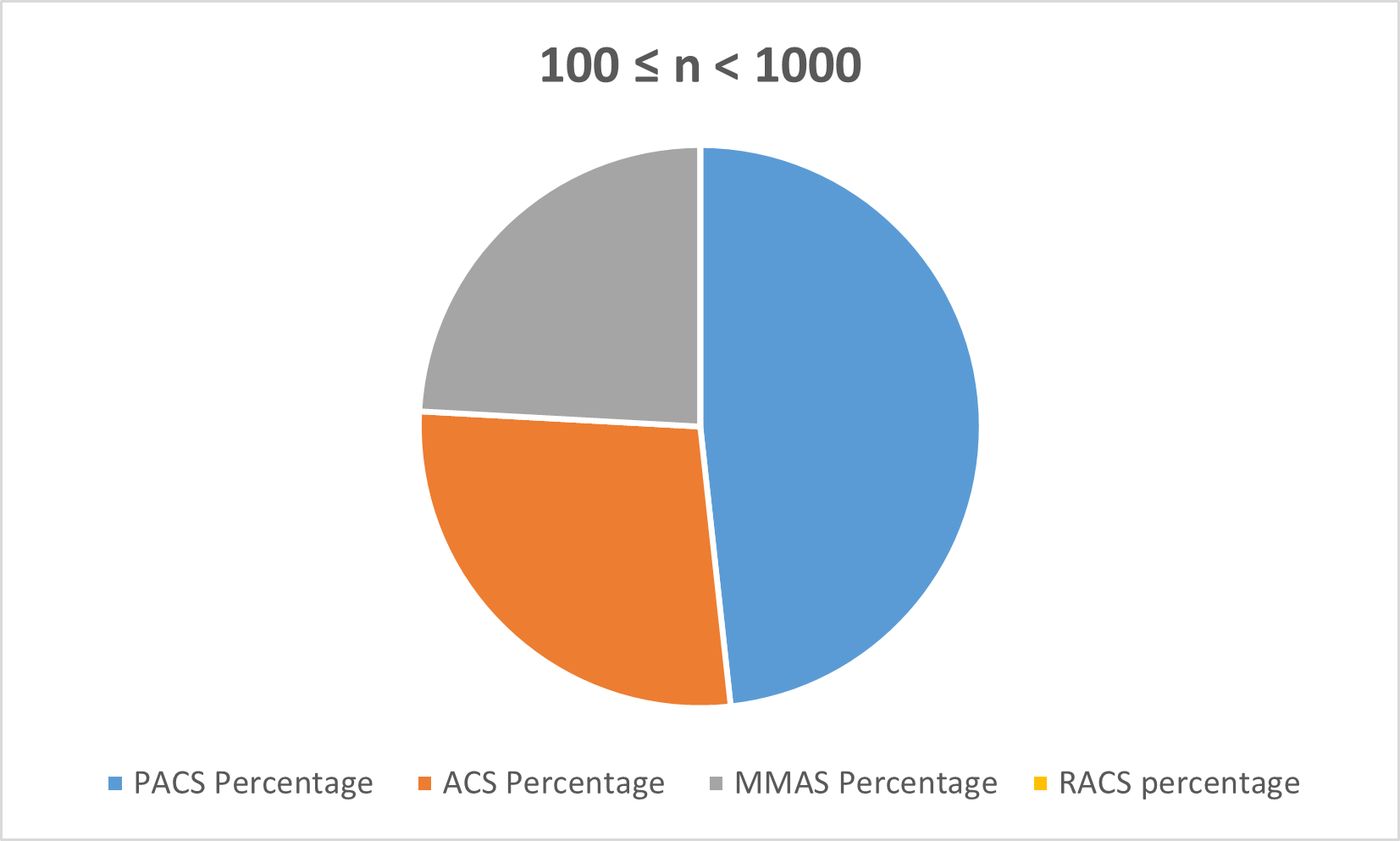}
        \caption{$100 \leq n < 1000$ Chart}
        \label{fig:sub2}
    \end{subfigure}
    \begin{subfigure}[b]{0.4\linewidth}
        \includegraphics[width=\linewidth]{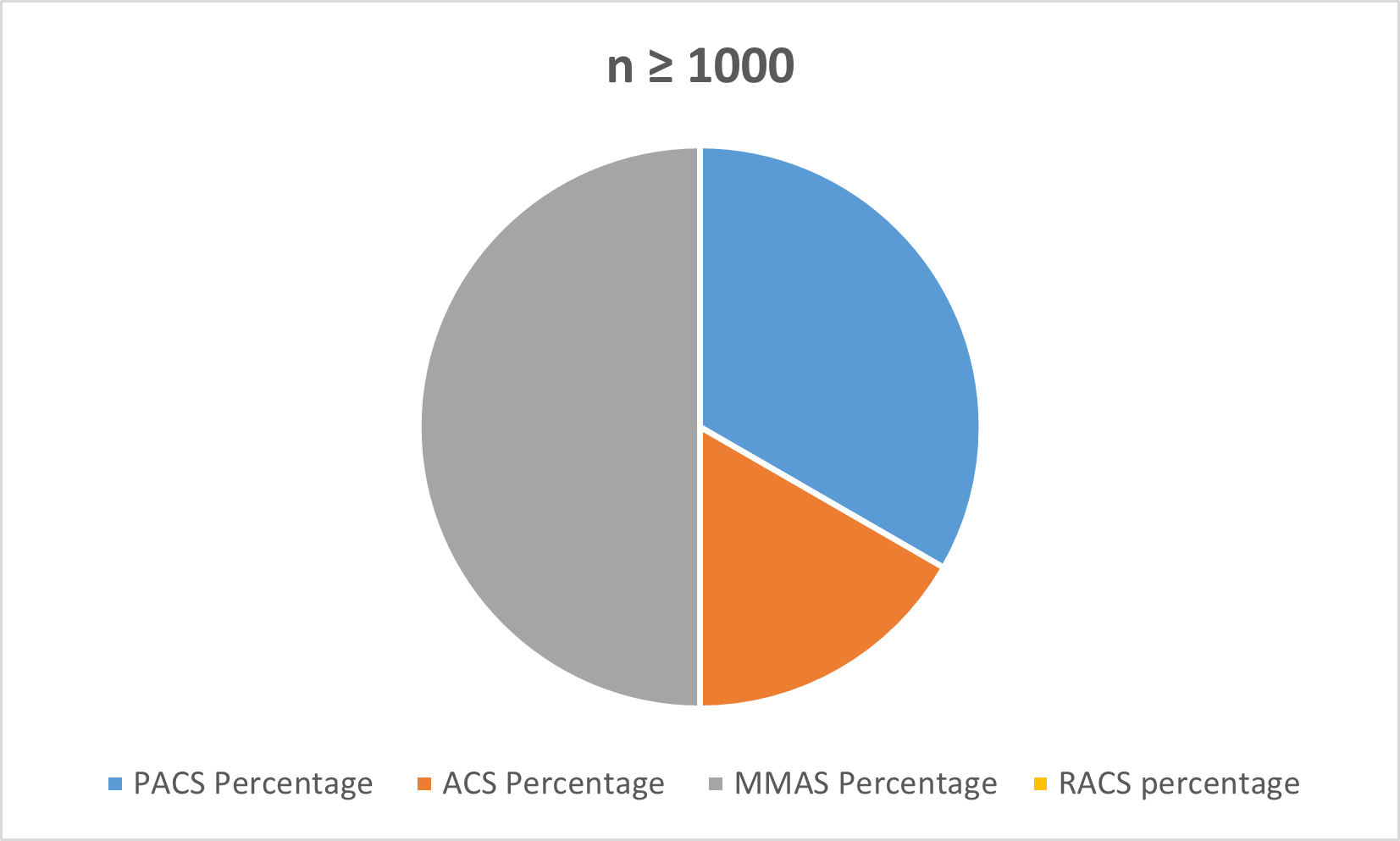}
        \caption{$n \geq 1000$ Chart}
        \label{fig:sub3}
    \end{subfigure}
    \caption{Comparison of Charts}
    \label{fig:comparison}
\end{figure}
\begin{table}[!ht]
\centering
\caption{$n < 100$ Table Data}
\label{tab:n100}
\begin{tabular}{|c|c|c|c|}
\hline
Dimensions & Best Distance & Best Algorithm & Memory consumption (MB) \\
\hline
48 & 65140.11 & PACS & 15.67188 \\
52 & 17124.66 & ACS & 15.59375 \\
14 & 31.37067 & ACS & 15.22656 \\
51 & 1003.574 & ACS & 15.39844 \\
76 & 1567.927 & MMAS & 15.125 \\
96 & 1786.259 & ACS & 11.59375 \\
\hline
\multicolumn{4}{|c|}{PACS Percentage: 16.66667\%} \\
\hline
\multicolumn{4}{|c|}{ACS Percentage: 66.66667\%} \\
\hline
\multicolumn{4}{|c|}{MMAS Percentage: 16.66667\%} \\
\hline
\multicolumn{4}{|c|}{ASRank percentage: 0\%} \\
\hline
\end{tabular}
\end{table}
\clearpage

\begin{table}[!ht]
\centering
\caption{$100 \leq n < 1000$ Table Data}
\label{tab:n1000}
\begin{tabular}{|c|c|c|c|}
\hline
Dimensions & Best Distance & Best Algorithm & Memory consumption (MB)\\
\hline
280 & 19233.58 & ACS & 15.00391 \\
535 & 17590.66 & PACS & 15.63281 \\
532 & 841740.3 & ACS & 15.08203 \\
127 & 403724.9 & ACS & 15.61719 \\
130 & 25464.67 & PACS & 15.63672 \\
150 & 31683.15 & ACS & 15.58594 \\
198 & 67667.7 & PACS & 11.47266 \\
493 & 261093.2 & PACS & 10.86719 \\
657 & 519156.7 & MMAS & 11.07813 \\
101 & 2014.44 & MMAS & 11.48438 \\
417 & 119981.2 & MMAS & 11.07813 \\
262 & 15738.57 & PACS & 15.60156 \\
137 & 3065.222 & PACS & 15.38672 \\
202 & 1964.933 & MMAS & 11.46484 \\
229 & 7754.671 & ACS & 11.27344 \\
431 & 12914.53 & MMAS & 11.84766 \\
666 & 30378.96 & PACS & 11.5625 \\
100 & 83566.46 & ACS & 12.20703 \\
150 & 132240.4 & PACS & 11.97656 \\
200 & 170933.6 & ACS & 12.21875 \\
100 & 80077.98 & PACS & 11.99219 \\
150 & 127134.7 & MMAS & 11.875 \\
200 & 172624.2 & PACS & 11.61328 \\
100 & 80364.69 & ACS & 11.51563 \\
100 & 80692.43 & PACS & 11.51953 \\
100 & 83669.4 & PACS & 11.96484 \\
105 & 55941 & MMAS & 15.54688 \\
318 & 325578.5 & PACS & 15.65234 \\
318 & 331943 & PACS & 11.87109 \\
\hline
\multicolumn{4}{|c|}{PACS Percentage: 48.27586207\%} \\
\hline
\multicolumn{4}{|c|}{ACS Percentage: 27.5862069\%} \\
\hline
\multicolumn{4}{|c|}{MMAS Percentage: 24.13793103\%} \\
\hline
\multicolumn{4}{|c|}{ASRank percentage: 0\%} \\
\hline
\end{tabular}
\end{table}
\clearpage

\begin{table}[!ht]
\centering
\caption{$n > 1000$ Table Data}
\label{tab:n1000p}
\begin{tabular}{|c|c|c|c|}
\hline
Dimensions & Best Distance & Best Algorithm & Memory consumption (MB)\\
\hline
1291 & 1035099 & ACS & 15.33203 \\
1655 & 1285758 & MMAS & 9.472656 \\
2103 & 277927.6 & MMAS & 12.63281 \\
1000 & 2.84E+08 & MMAS & 15.19531 \\
1400 & 344063.2 & PACS & 12.70313 \\
1577 & 663057.1 & PACS & 11.69922 \\
\hline
\multicolumn{4}{|c|}{PACS Percentage: 33.33333333\%} \\
\hline
\multicolumn{4}{|c|}{ACS Percentage: 16.66666667\%} \\
\hline
\multicolumn{4}{|c|}{MMAS Percentage: 50\%} \\
\hline
\multicolumn{4}{|c|}{ASRank percentage: 0\%} \\
\hline
\end{tabular}
\end{table}

\subsection{Key Findings and Insights}

\begin{enumerate}
\item \textbf{ACS: The Sprinter for Smaller Problems:} ACS consistently outperformed other algorithms on smaller TSP instances, demonstrating its ability to converge quickly and effectively exploit promising solutions in less complex search spaces. This strength likely stems from its aggressive action choice rule and local pheromone updates, which favor rapid intensification of the search. However, ACS's performance became less consistent with larger instances, suggesting a tendency toward premature convergence and limitations in exploring diverse solutions.

\item \textbf{PACS: The Adaptable Choice for Medium-Scale Complexity:} PACS emerged as a strong contender for medium-sized TSP instances, showcasing its adaptability to increasing problem complexity. Its pop-ulation-based approach, which implicitly represents pheromone information, appears to provide a better balance between exploration and exploitation compared to algorithms that rely on the explicit pheromone matrix.

\item \textbf{MMAS: The Marathon Runner for Large-Scale Challenges:} \\ MMAS consistently delivered competitive results across all problem scales, particularly excelling in larger and more challenging TSP instances. Its pheromone trail limits and selective pheromone update rule seem to play a crucial role in mitigating stagnation, promoting wider exploration, and avoiding premature convergence to local optima. Although MMAS may exhibit slower initial convergence, its ability to continuously improve solutions makes it well-suited for complex TSP scenarios.

\item \textbf{ASRank: Limited Success and the Need for Diversification:} ASRank struggled to outperform the other algorithms, indicating limitations in its ranking and elitism mechanism. While this approach may accelerate initial convergence, it appears less effective in the long run, potentially leading to stagnation or premature convergence. This finding highlights the need for incorporating diversification strategies or exploring alternative pheromone update rules to enhance ASRank's performance.
\end{enumerate}

The study also confirmed that:

\begin{itemize}
\item \textbf{Problem Size Matters:} The performance of all algorithms generally degrades with increasing problem size, exhibiting longer execution times and less consistent solution quality. This underscores the challenge of scaling ACO algorithms for very large TSP instances and emphasizes the need for ongoing research on more efficient node selection procedures, optimized memory management, and parallel or distributed computing techniques.

\item \textbf{Beyond Size: The Influence of Instance Type and Structure:} The performance of ACO algorithms can be significantly impacted by the type of TSP instance (symmetric or asymmetric) and the structure of the search space. ACS, for example, performed well on symmetric instances with a clear structure but struggled on asymmetric instances. Algorithms like MMAS and PACS, with more flexible exploration mechanisms, demonstrated greater robustness across different TSP instance types.
\end{itemize}

\section{Conclusion and future work}

This research undertook a comparative analysis of four prominent Ant Colony Optimization (ACO) algorithms - ASRank, MMAS, ACS, and PACS - applied to the Traveling Salesman Problem (TSP). By examining their performance across a diverse set of TSP instances, we aimed to understand the strengths, weaknesses, and scalability of each algorithm. Our findings revealed a nuanced landscape of performance, highlighting the importance of selecting the right ACO algorithm based on the specific problem characteristics and available computational resources. 

Several avenues for future research emerge from this study:

\begin{enumerate}
\item \textbf{Memory Efficiency:} A deeper investigation of the memory usage patterns of ACO algorithms throughout their execution is needed to identify potential bottlenecks and develop more efficient memory management strategies for large-scale TSP instances.
\item \textbf{Hybridization and Parallel Computing:} Exploring the hybridization of ACO algorithms with other metaheuristics or local search techniques, as well as implementing algorithms in parallel and distributed computing environments, could enhance their scalability and performance for massive TSP instances.
\item \textbf{Dynamic TSP:} Adapting and evaluating the performance of ACO algorithms on dynamic TSP instances, where the problem characteristics change over time, presents a promising area for future research.
\end{enumerate}

\end{document}